\documentclass{article}

\usepackage{arxiv}

\usepackage[utf8]{inputenc} 
\usepackage{amsmath}        
\usepackage[T1]{fontenc}    
\usepackage{hyperref}       
\usepackage{url}            
\usepackage{booktabs}       
\usepackage{amsfonts}       
\usepackage{nicefrac}       
\usepackage{microtype}      
\usepackage{lipsum}
\usepackage{subcaption} 
\usepackage{graphicx}
\graphicspath{ {./images/} }

\title{Scientific machine learning in ecological systems: A study on the predator-prey dynamics}

\date{September 20, 2024} 

\author{
 Ranabir Devgupta \\
 RCC Institute of Information Technology\\\
  \texttt{rdgupta10@gmail.com} \\
	\And
	{Raj Abhijit Dandekar} \\
	Vizuara AI Labs\\
	\texttt{raj@vizuara.com} \\
 \And
	{Rajat Dandekar} \\
	Vizuara AI Labs\\
	\texttt{rajatdandekar@vizuara.com} \\
 \And
	{Sreedath Panat} \\
	Vizuara AI Labs\\
	\texttt{sreedath@vizuara.com} \\
}

\begin{document}
\maketitle
\vspace{-1em} 

\begin{center}
    PREPRINT 
\end{center}
\begin{abstract}
In this study, we apply two pillars of Scientific Machine Learning: Neural Ordinary Differential Equations (Neural ODEs) and Universal Differential Equations (UDEs) to the Lotka-Volterra Predator-Prey Model, a fundamental ecological model describing the dynamic interactions between predator and prey populations. The Lotka-Volterra model is critical for understanding ecological dynamics, population control, and species interactions, as it is represented by a system of differential equations. 
In this work, we aim to uncover the underlying differential equations without prior knowledge of the system, relying solely on training data and neural networks. Using robust modeling in the Julia programming language, we demonstrate that both Neural ODEs and UDEs can be effectively utilized for prediction and forecasting of the Lotka-Volterra system. More importantly, we introduce the "forecasting breakdown point" – the time at which forecasting fails for both Neural ODEs and UDEs. We observe how UDEs outperform Neural ODEs by effectively recovering the underlying dynamics and achieving accurate forecasting with significantly less training data. 
Additionally, we introduce Gaussian noise of varying magnitudes (from mild to high) to simulate real-world data perturbations and show that UDEs exhibit superior robustness, effectively recovering the underlying dynamics even in the presence of noisy data, while Neural ODEs struggle with high levels of noise.
Through extensive hyperparameter optimization, we offer insights into neural network architectures, activation functions, and optimizers that yield the best results. This study opens the door to applying Scientific Machine Learning frameworks for forecasting tasks across a wide range of ecological and scientific domains.
\end{abstract}


\section{Introduction}

Scientific Machine Learning (SciML) is an emerging interdisciplinary field that integrates the strengths of traditional scientific models with the flexibility of machine learning techniques, and it has seen successful applications across various domains such as epidemiology, gene regulation, quantum mechanics, and fluid dynamics \cite{ref1, PINNeqn, rackauckas2020universal}. The core of SciML lies in combining the structure and interpretability of physical models, such as ordinary and partial differential equations (ODEs/PDEs) \cite{rackauckas2017differentialequations,ma2021comparison}, with the representational power of neural networks.

Two primary approaches define the landscape of Scientific Machine Learning:
\begin{itemize}
    \item \textbf{Neural Ordinary Differential Equations (Neural ODEs)}: This method replaces the traditional ODE/PDE system entirely with neural networks. By backpropagating through the system, we can optimize the network's parameters and approximate the dynamics of the system. The neural network is used to model the continuous-time dynamics of a system. Neural ODEs treat the hidden layers of a neural network as continuous transformations, modeled by a differential equation, rather than discrete layers as in conventional neural networks. This formulation allows the use of ODE solvers to propagate the system state forward in time. \cite{ref1, rackauckas2020universal}.
    The general form of a Neural ODE is given as:

    \begin{equation}
        \frac{du(t)}{dt} = f_\theta(u(t), t)
    \end{equation}
    
    where \( u(t) \) represents the system state at time \( t \), and \( f_\theta \) is a neural network parameterized by \( \theta \). In this formulation, the neural network learns the time derivatives (i.e., the dynamics) of the state directly from the data, without requiring a predefined model for the system's evolution.

    The continuous transformation of the state variable is modeled as:
    
    \begin{equation}
        u(t_1) = u(t_0) + \int_{t_0}^{t_1} f_\theta(u(t), t) \, dt
    \end{equation}
    
    where \( u(t_0) \) is the initial state and \( u(t_1) \) is the state at time \( t_1 \). The system state can be propagated forward by solving this integral using numerical ODE solvers such as the Runge-Kutta methods (e.g., Tsit5) or adaptive solvers. The objective is to minimize the difference between the predicted and observed state trajectories by optimizing the parameters \( \theta \) of the neural network.
    
    The training process of Neural ODEs typically involves backpropagating through the ODE solver. However, directly backpropagating through the entire integration process can be computationally expensive. To address this, the \textit{adjoint sensitivity method} is commonly employed. The adjoint method computes gradients with respect to the neural network's parameters \( \theta \) by solving an additional ODE backwards in time, thus allowing efficient memory management during training. This method is summarized as follows:

    \begin{equation}
    \frac{dL}{d\theta} = \int_{t_1}^{t_0} a(t)^T \frac{\partial f_\theta(u(t), t)}{\partial \theta} \, dt
    \end{equation}
    
    where \( a(t) \) is the adjoint variable, which captures how the loss \( L \) changes with respect to the state variables.

    \item \textbf{Universal Differential Equations (UDEs)}: Instead of substituting the entire equation, UDEs replace specific terms of ODEs/PDEs with neural networks, allowing the discovery of unknown components or missing dynamics in existing models \cite{rackauckas2020universal}. This approach is particularly useful when some physical laws are known but others remain uncertain. While Neural ODEs replace the entire system with a neural network, UDEs retain certain known components of the differential equation and augment the unknown or uncertain parts with neural networks. This method allows the system to capture known physics while also learning unknown dynamics from the data.

    The general form of a UDE is:
    
    \begin{equation}
        \frac{du(t)}{dt} = f_{\text{known}}(u(t), t) + f_{\text{NN}}(u(t), t; \theta)
    \end{equation}
    
    where \( f_{\text{known}} \) represents the known part of the differential equation derived from physical laws, and \( f_{\text{NN}} \) is a neural network with parameters \( \theta \) that models the unknown or unmodeled dynamics. The neural network's role is to fill in the gaps where the traditional model is insufficient or where the underlying mechanisms are unknown.
    
    
    In practice, the UDE takes the form:
    
    \begin{equation}
        \frac{du(t)}{dt} = f_{\text{known}}(u(t), t) + \sum_{i=1}^{m} \text{NN}_i(u(t), t; \theta_i)
    \end{equation}
    
    where the neural networks \( \text{NN}_i \) are used to model specific unknown terms or interactions in the system. Each neural network takes the system state \( u(t) \) and possibly time \( t \) as inputs, and outputs the corresponding missing term in the equation. The sum allows for multiple neural networks to model different parts of the dynamics.

\end{itemize}

Despite the significant progress of SciML in various scientific domains like physics, chemistry, finance, ecological models etc \cite{rackauckas2020universal, dandekar2020quantifying, toth2020hamiltonian, rackauckas2019scalable,kidger2020neural}, a  comprehensive studies on forecasting accuracy, model breakdown points, and robust hyperparameter optimization have yet to be fully addressed which upon here .

This work addresses the following key questions:
\begin{itemize}
    \item Can we leverage UDEs to learn the interaction terms in a predator-prey system by replacing traditional terms with neural networks?
    \item How do predictions from Neural ODEs compare with those from UDEs in ecological modeling?
    \item Can Neural ODE/UDE recover interaction terms from less training data - if so, then what is potential forecast breakdown point for both
    
    \item Is forecasting with UDEs more effective than with Neural ODEs?
    
\end{itemize}

To answer these questions, we utilize the Lotka-Volterra predator-prey model, a foundational system in ecology, as a case study. Using this model, we aim to demonstrate how UDEs can enhance our understanding of population dynamics by learning unknown interaction terms, while comparing the results with traditional Neural ODEs.

In this work, we use the powerful Scientific Machine Learning libraries provided by Julia \cite{rackauckas2017differentialequations, rackauckas2020universal} for efficient differential equation solving and neural network integration. Through extensive hyperparameter tuning and optimization, we explore how neural network architectures, activation functions, and optimization strategies affect the performance of both Neural ODEs and UDEs. We observe how UDEs can effectively recover underlying model by introducing Gaussian Noise to mimic real world noise data perturbations. Additionally, we introduce the concept of a "forecast breakdown point"— the moment beyond which predictive accuracy significantly deteriorates. This insight sheds light on the limitations of current SciML frameworks for long-term forecasting tasks.

The structure of the paper is as follows: we begin by detailing the methodology behind Neural ODEs and UDEs. We then present results from predictions and forecasts using both methods, followed by hyperparameter optimization outcomes. Lastly, we conclude with an analysis of the findings and discuss future directions for the application of Scientific ML in ecological modeling.

\section{Methodology}

In this work, we aim to model the Lotka-Volterra Predator-Prey system using two advanced approaches: \textbf{Neural Ordinary Differential Equations (Neural ODEs)} and \textbf{Universal Differential Equations (UDEs)}. These methods leverage the flexibility of neural networks \cite{hinton2012neural,goodfellow2016deep} to model or learn specific interaction terms within the system. 

The Lotka-Volterra equations, which describe the interaction between prey and predator populations, are given by the system of first-order ordinary differential equations (ODEs):

\begin{equation}
    \frac{dx}{dt} = \alpha x - \beta x y
\end{equation}

\begin{equation}
    \frac{dy}{dt} = -\delta y + \gamma x y
\end{equation}

where \( x(t) \) represents the prey population, \( y(t) \) represents the predator population, and \( \alpha \), \( \beta \), \( \gamma \), and \( \delta \) are constant parameters that govern the interaction dynamics.

\subsection{Data Generation}

To generate synthetic data for training and testing, we set the parameters of the Lotka-Volterra system to \( \alpha = 1.5 \), \( \beta = 1.0 \), \( \gamma = 0.5 \), and \( \delta = 2.0 \), with initial conditions \( x(0) = 1.0 \) and \( y(0) = 1.0 \). We numerically solve the system over the time span \( t \in [0, 10] \) using the \textbf{Tsitouras 5/4 Runge-Kutta method (Tsit5)} with 101 equally spaced time points. [Refer Fig. 1(a)]

The solution provides time-series data for both prey and predator populations. For UDE, Gaussian noise with a standard deviation of \( \sigma = 0.05, 0.1, 0.3 \) was added to simulate real-world data perturbations. [Refer Fig. 1(b)]

\begin{figure}[h]
    \centering
    \begin{subfigure}{0.45\textwidth}
        \centering
        \includegraphics[width=\linewidth]{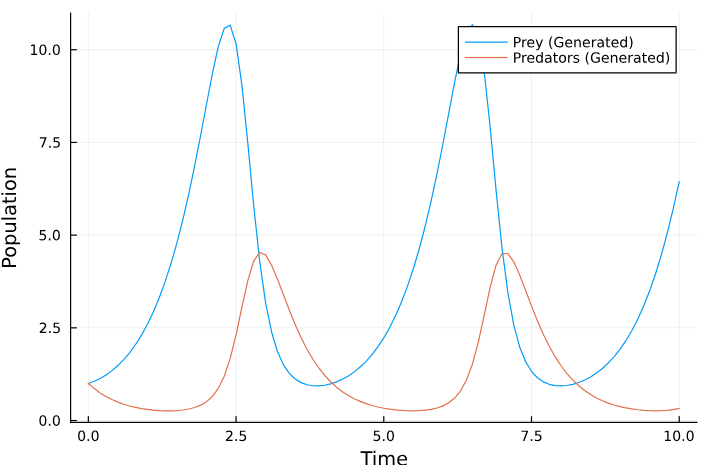}  
        \caption{Generated Synthetic Data}
        \label{fig:image1}
    \end{subfigure}
    \hfill
    \begin{subfigure}{0.45\textwidth}
        \centering
        \includegraphics[width=\linewidth]{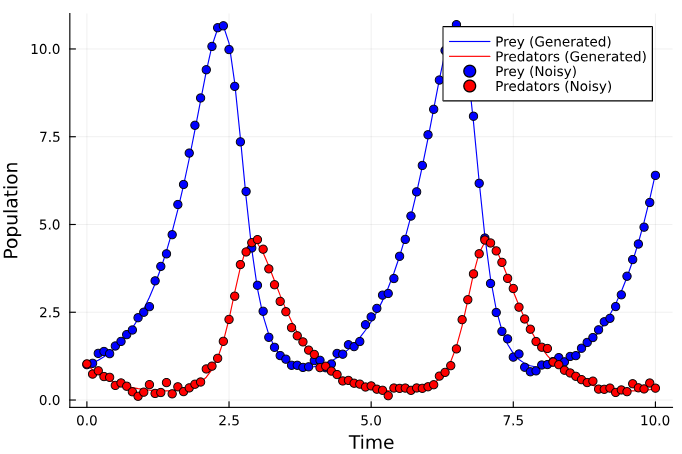}  
        \caption{Generated Synthetic Data with Gaussian noise of standard deviation 0.1}
        \label{fig:image2}
    \end{subfigure}
    \caption{Generated Lotka-Volterra Synthetic Data}
    \label{fig:sidebyside}
\end{figure}

\subsection{Neural Ordinary Differential Equation (Neural ODE)}

In the Neural ODE approach, we replace the entire right-hand side of the Lotka-Volterra system with a neural network that learns the underlying dynamics directly from data. The neural network architecture consists of fully connected layers using radial basis functions (RBFs) as activation functions. The structure is as follows:
\begin{itemize}
    \item The input layer takes the 2-dimensional vector \( [x, y] \).
    \item Three hidden layers, each with 100 neurons, and the RBF activation function.
    \item The output layer produces a 2-dimensional vector \( [\frac{dx}{dt}, \frac{dy}{dt}] \).
\end{itemize}

The loss function for the Neural ODE is RMSE defined as the sum of squared differences between the predicted and actual prey and predator populations:

\begin{equation}
    \mathcal{L}(\theta) = \sum_{i=1}^{n} \left( (x_i^{\text{true}} - \hat{x}_i(\theta))^2 + (y_i^{\text{true}} - \hat{y}_i(\theta))^2 \right)
\end{equation}

where \( x_i^{\text{true}} \) and \( y_i^{\text{true}} \) are the true prey and predator data, and \( \hat{x}_i(\theta) \) and \( \hat{y}_i(\theta) \) are the predicted values from the neural network.

\subsection{Universal Differential Equation (UDE)}

In the UDE approach, we retain certain known components of the Lotka-Volterra system (such as prey growth and predator death rates) and replace the interaction terms \( \beta x y \) and \( \gamma x y \) with neural networks. The UDE system is formulated as:

\begin{equation}
    \frac{dx}{dt} = \alpha x - NN_1(x, y; \theta_1)
\end{equation}

\begin{equation}
    \frac{dy}{dt} = -\delta y + NN_2(x, y; \theta_2)
\end{equation}

where \( NN_1 \) and \( NN_2 \) are neural networks that model the unknown interaction terms. Each network takes \( [x, y] \) as input and outputs a scalar. The training procedure for UDE is similar to that of the Neural ODE, where the loss function is minimized by comparing predicted and actual prey and predator populations.

For UDE the neural network parameters \( \theta_1 \) and \( \theta_2 \) are initialized randomly, and the network is trained by minimizing the difference between synthetic and predicted data.
By introducing the Gaussian noise the loss is calculated by of  difference noisy synthetic data and the predicted populations from the UDE model. The loss function is defined as the sum of squared errors between the noisy prey and predator data and the UDE-predicted values:

\begin{equation}
    \mathcal{L}(\theta) = \sum_{i=1}^{n} \left( (x_i^{\text{noisy}} - \hat{x}_i(\theta))^2 + (y_i^{\text{noisy}} - \hat{y}_i(\theta))^2 \right)
\end{equation}

where \( x_i^{\text{noisy}} \) and \( y_i^{\text{noisy}} \) are the noisy prey and predator data, and \( \hat{x}_i(\theta) \) and \( \hat{y}_i(\theta) \) are the predicted values from the UDE model.

\subsection{Model Training and Optimization}

\subsubsection{Neural Network Architecture and Performance}

The architecture of the neural networks plays a critical role in how well both the Neural ODE and UDE models perform. While both models use neural networks to learn the underlying dynamics of the Lotka-Volterra system, their structural design and depth directly influence the model’s ability to generalize and make accurate predictions.

\subsubsection{Neural ODE Architecture}

The neural network architecture in the Neural ODE model is relatively deep, comprising three hidden layers with 100 neurons each. Figure 5 shows the impact of different number of hidden units vs Loss. 
We will discuss this more on section 2.5 in hyperparameters

\subsubsection{UDE Architecture and the Advantage of a Shallow Network}

In contrast, the UDE model leverages a much shallower neural network architecture. Specifically, two neural networks with three hidden layers and only 10 neurons per layer are used. The activation function chosen is the Rectified Linear Unit (ReLU), which is computationally efficient and helps in avoiding vanishing gradient issues commonly seen with deeper networks \cite{nair2010rectified,glorot2011deep,maas2013rectifier}.

The shallow architecture of the UDE has several advantages:

\begin{itemize}
    \item \textbf{Faster Training:} The shallow architecture results in fewer parameters to optimize, which leads to faster convergence during training. This reduced complexity allows the UDE model to reach accurate predictions with fewer training iterations compared to the deeper Neural ODE model.
    \item \textbf{Better Generalization:} While deep networks have high capacity, they are more prone to overfitting, especially when trained on noisy or limited data. The shallow network used in the UDE model strikes a balance between model capacity and generalization, making it less likely to overfit to the training data.
\end{itemize}

\textbf{The key advantage of using a shallow architecture in UDE comes from the fact that the model is not tasked with learning the entire dynamics from scratch.} Instead, it only needs to learn the residual terms (e.g., interaction terms like $\beta$xy and $\gamma$xy in the Lotka-Volterra system). This partial learning task allows the model to remain simple, robust, and computationally efficient. In contrast, the Neural ODE approach requires a more complex architecture as it must approximate the entire system behavior, leading to the need for a deeper network.

\subsubsection{Model Optimization}
Both the Neural ODE and Universal Differential Equation (UDE) models were trained using a two-phase optimization strategy. This combined approach leverages the strengths of different optimizers to first rapidly converge on an approximate solution, followed by a more refined optimization to reach the optimal parameters.

\subsubsection{Neural ODE Optimization}

For the Neural ODE model, the optimization was performed in two phases [Refer Fig 2(a)]:

\begin{itemize}
    \item \textbf{Adam Optimization:} The Adam optimizer \cite{kingma2014adam,wilson2017marginal,choi2019empirical,loshchilov2017fixing} was employed for 400 iterations with a learning rate of 0.001. Adam was chosen for its adaptability and ability to handle sparse gradients.
    \item \textbf{BFGS Optimization:} After the initial phase with Adam, the parameters were further refined using the BFGS optimization \cite{powell2009bfgs,liu1989limited,shanno1970conditioning} algorithm for 100 iterations. BFGS, being a quasi-Newton method, allows for fine-tuning the parameters to a local minimum with greater precision.
\end{itemize}

\subsubsection{UDE Optimization}

Similarly, the UDE model training also used a two-phase optimization process [Refer Fig 2(b)]:

\begin{itemize}
    \item \textbf{Adam Optimization:} The Adam optimizer was applied for 20,000 iterations with a learning rate of 0.001. Given its effectiveness in adjusting the learning rate dynamically, Adam was well-suited for the UDE model, especially in handling noisy data.
    \item \textbf{RMSProp Optimization:} After the Adam phase, RMSProp \cite{tieleman2012rmsprop} was used to further optimize the model for an additional 20,000 iterations. With a learning rate of 0.001, momentum \( \rho = 0.9 \), and \( \epsilon = 1e{-8} \), RMSProp helped fine-tune the parameters, particularly in non-stationary environments like the UDE model.
\end{itemize}

\begin{figure}[h]
    \centering
    \begin{subfigure}{0.45\textwidth}
        \centering
        \includegraphics[width=\linewidth]{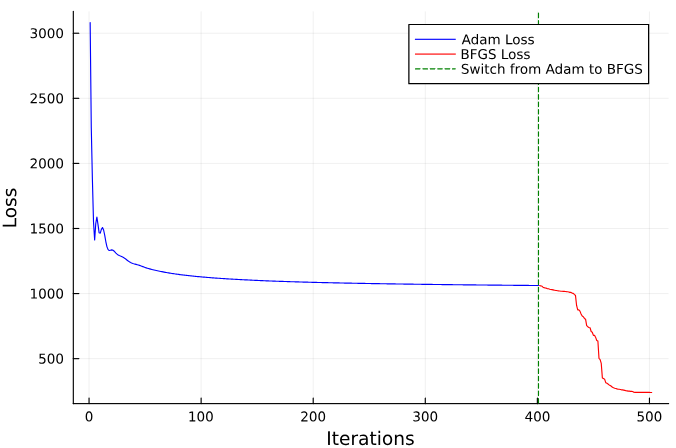}  
        \caption{Neural ODE Loss Variation with Optimizer Adam and BFGS}
        \label{fig:loss_variation_switch_ODE}
    \end{subfigure}
    \hfill
    \begin{subfigure}{0.45\textwidth}
        \centering
        \includegraphics[width=\linewidth]{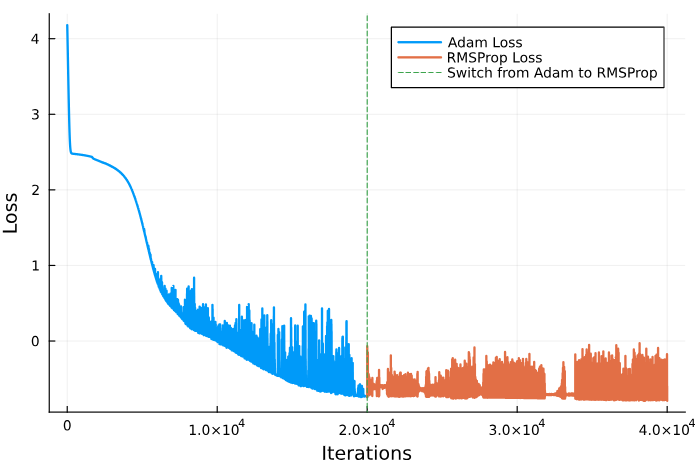}  
        \caption{UDE Loss Variation (Lograithmic Scale of base 10) with Optimizer Adam and RMSProp}
        \label{fig:loss_variation_switch}
    \end{subfigure}
    \caption{Loss variation during training for both ODE and UDE models.}
    \label{fig:loss_variation_optimizers}
\end{figure}

\subsection{Hyperparameters}

\begin{figure}[h!]
    \centering
    \begin{subfigure}[b]{0.45\textwidth}
        \centering
        \includegraphics[width=\textwidth]{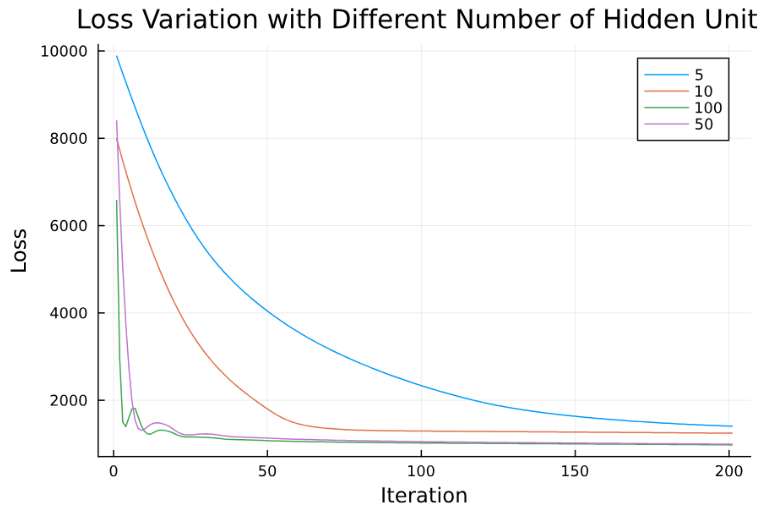}
        \caption{Neural ODE: Loss vs Hidden Units}
        \label{fig:neural_ode_hidden_units}
    \end{subfigure}
    \hfill
    \begin{subfigure}[b]{0.45\textwidth}
        \centering
        \includegraphics[width=\textwidth]{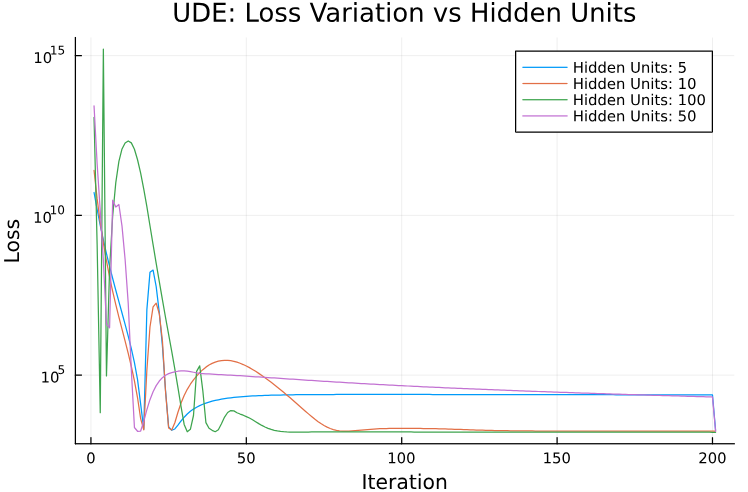}
        \caption{UDE: Loss vs Hidden Units}
        \label{fig:ude_hidden_units}
    \end{subfigure}

    \begin{subfigure}[b]{0.45\textwidth}
        \centering
        \includegraphics[width=\textwidth]{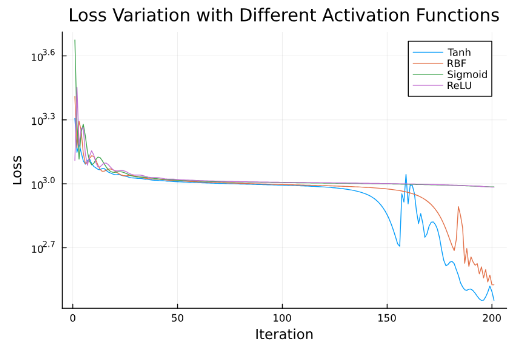}
        \caption{Neural ODE: Loss vs Activation Function}
        \label{fig:neural_ode_activation_function}
    \end{subfigure}
    \hfill
    \begin{subfigure}[b]{0.45\textwidth}
        \centering
        \includegraphics[width=\textwidth]{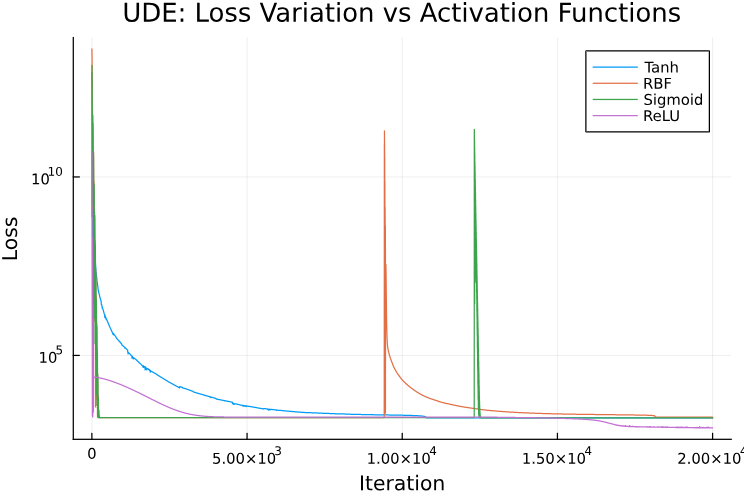}
        \caption{UDE: Loss vs Activation Function}
        \label{fig:ude_activation_function}
    \end{subfigure}

    \begin{subfigure}[b]{0.45\textwidth}
        \centering
        \includegraphics[width=\textwidth]{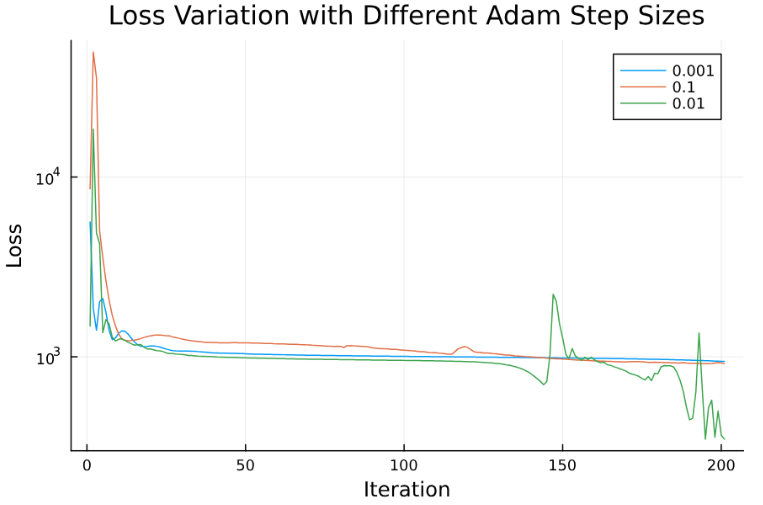}
        \caption{Neural ODE: Loss vs Step Size}
        \label{fig:neural_ode_step_size}
    \end{subfigure}
    \hfill
    \begin{subfigure}[b]{0.45\textwidth}
        \centering
        \includegraphics[width=\textwidth]{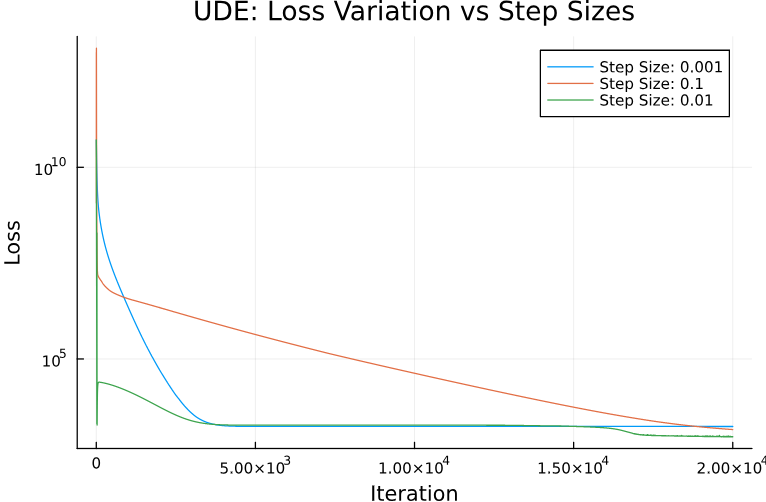}
        \caption{UDE: Loss vs Step Size}
        \label{fig:ude_step_size}
    \end{subfigure}
    
    \caption{Comparison of Loss Variation with Hidden Units, Activation Functions, and Step Size between Neural ODE and UDE. The first row shows loss variation with hidden units, the second row with activation functions, and the third row with step size.}
    \label{fig:comparison_loss_neural_ode_ude}
\end{figure}

Hyperparameter tuning is a critical aspect of the model training process. For both Neural ODE and UDE models, specific hyperparameters have a significant impact on model performance, stability, and convergence Refer Figure \ref{fig:comparison_loss_neural_ode_ude}. In this section, we will explore the key hyperparameters used for each model and how they influence the training process.

\subsubsection{Neural ODE Hyperparameters}

The Neural ODE model uses a deeper architecture and thus requires careful tuning of several key hyperparameters to ensure stable and efficient training. The main hyperparameters considered for the Neural ODE are:

\begin{itemize}
    \item \textbf{Learning Rate:} A learning rate of 0.01 was used with the Adam optimizer. The learning rate controls the step size at each iteration while moving towards a minimum in the loss function. For Neural ODEs, a learning rate that is too high can lead to oscillations or divergence, while a very small learning rate can result in slow convergence. The value of 0.01 was chosen as a compromise between these extremes, providing stable convergence without overshooting. 
    
    \item \textbf{Number of Layers and Neurons:} The neural network used in the Neural ODE has 3 hidden layers, each with 100 neurons [Refer Figure 4]. This deep architecture is essential to capture the complex dynamics of the Lotka-Volterra system, but it also introduces a large number of trainable parameters. The depth and width of the network significantly impact the expressiveness of the model. However, deeper networks often require more computational resources and are prone to overfitting, which is why regularization techniques and careful monitoring of the validation loss are critical.

    \item \textbf{Activation Function:} The Radial Basis Function (RBF) was chosen as the activation function for the hidden layers. RBFs are particularly effective in modeling complex, non-linear interactions as they provide a smooth, localized response. However, their complexity adds a computational burden and requires more careful initialization of weights and biases to avoid numerical instability during training.

    \item \textbf{Optimizers:} A two-phase optimization strategy was used:
        \begin{itemize}
            \item \textbf{Adam Optimizer:} Adam, with a learning rate of 0.001, was employed for the first 500 iterations. Adam’s adaptability to the varying learning rates during training makes it an excellent choice for initial optimization, especially for complex models with noisy gradients. Adam's momentum terms allow it to navigate complex loss surfaces and escape local minima more efficiently.
            \item \textbf{BFGS Optimizer:} After the Adam optimization phase, BFGS (Broyden-Fletcher-Goldfarb-Shanno) was used for 100 iterations to refine the parameters. BFGS is a quasi-Newton method and excels at fine-tuning the solution found by Adam by efficiently finding the local minima in smooth and continuous optimization landscapes. Its reliance on second-order approximations of the Hessian matrix allows for more precise parameter adjustments, making it suitable for final parameter refinement.
        \end{itemize}

    \item \textbf{Epochs/Iterations:} In the initial phase, 500 iterations were conducted with the Adam optimizer. This was followed by 100 iterations using BFGS. The number of iterations was chosen to ensure a good balance between convergence speed and model performance.
    
    \item \textbf{Weight Initialization:} Proper weight initialization is crucial, especially for deep networks. Random initialization using default settings in the Lux package was employed, ensuring that the initial values were not too large (which can lead to exploding gradients) or too small (which can result in vanishing gradients).
\end{itemize}

\subsubsection{UDE Hyperparameters}

For the UDE model, the neural network architecture is shallower, and the focus is on learning specific interaction terms, which allows for a more streamlined set of hyperparameters. The key hyperparameters for the UDE model include:

\begin{itemize}
    \item \textbf{Learning Rate:} Similar to the Neural ODE, a learning rate of 0.001 was used with the Adam optimizer. However, given the shallow architecture of the UDE, the model is less sensitive to the learning rate compared to the deeper Neural ODE model. The learning rate is crucial in the early stages of training, ensuring that the UDE model can rapidly converge to a reasonable approximation of the predator-prey interactions.
    
    \item \textbf{Number of Layers and Neurons:} The UDE model employs two separate neural networks with 3 hidden layers and 10 neurons each. This shallow architecture significantly reduces the number of trainable parameters, making the training process more efficient. The reduced number of neurons is sufficient for learning the residual terms (like $\beta$xy and $\gamma$xy in the Lotka-Volterra system), allowing the UDE model to generalize better and avoid overfitting. The small architecture also ensures faster convergence and reduces the computational load.

    \item \textbf{Activation Function:} The ReLU (Rectified Linear Unit) activation function was used in the UDE model. ReLU is computationally efficient and helps in avoiding the vanishing gradient problem, which can be an issue in deep networks. In a shallow network like UDE, ReLU provides a good balance between simplicity and effectiveness, allowing the model to learn complex interactions while maintaining numerical stability.

    \item \textbf{Optimizers:} A two-phase optimization strategy was also employed for the UDE model:
        \begin{itemize}
            \item \textbf{Adam Optimizer:} Adam, with a learning rate of 0.001, was used for 20,000 iterations. Given the lower number of parameters in the UDE model, Adam's dynamic adjustment of learning rates enabled rapid convergence to a good solution.
            \item \textbf{RMSProp Optimizer:} After the Adam phase, RMSProp was applied for an additional 5,000 iterations. RMSProp is particularly useful for non-stationary environments, as it adjusts the learning rate based on recent gradients, helping to avoid overshooting in the later stages of training. With RMSProp, we used a learning rate of 0.001, momentum $\rho = 0.9$, and $\epsilon = 1 \times 10^{-8}$ to stabilize training and ensure smooth convergence.
        \end{itemize}

    \item \textbf{Epochs/Iterations:} The UDE model underwent 20,000 iterations using Adam, followed by 5,000 iterations with RMSProp. The larger number of iterations is justified by the shallow network architecture, which allows for more stable and less computationally expensive optimization steps over a longer period.

    \item \textbf{Weight Initialization:} The Lux package’s default initialization strategy was used, with care taken to avoid too large or too small weights that could destabilize training. The small number of parameters in the UDE model made this process more straightforward and less prone to initialization issues compared to the Neural ODE.
\end{itemize}

\subsubsection{Comparison of Hyperparameters between Neural ODE and UDE}

The distinct architectures and roles of Neural ODE and UDE models explain why RBF worked better in Neural ODE, while ReLU was more effective in UDE.

\textbf{Neural ODE (RBF):}  
In the Neural ODE model, the neural network directly learns the continuous dynamics of the system, approximating the behavior of the entire Lotka-Volterra model over time. The RBF activation function excels in this setting because of its smooth, localized response. This allows it to model complex, non-linear, and continuous population dynamics with greater precision. Since differential equations require smooth transitions for stability and accurate prediction, RBF is a natural fit. Its ability to capture smooth, non-linear changes helps Neural ODE achieve better approximation and forecasting performance in this context.

\textbf{UDE (ReLU):}  
The UDE model, on the other hand, only learns residual interaction terms (e.g., $\beta xy$ and $\gamma  xy$) rather than the entire system's dynamics. Since UDE is not responsible for modeling the full continuous dynamics but instead corrects or complements an existing system (Lotka-Volterra), the simplicity and efficiency of ReLU are well-suited. ReLU’s piecewise linearity and computational efficiency make it effective for learning simpler, localized corrections to the system. The shallow architecture of UDE and the lower complexity of learning only residual terms allow ReLU to perform well without requiring the smoother transitions that RBF offers.

In summary, RBF’s smoothness is critical for Neural ODE’s task of learning continuous system dynamics, while ReLU’s simplicity and efficiency are ideal for UDE’s task of learning specific interaction corrections in a more straightforward, shallow architecture.

While both models share some common hyperparameters, such as the learning rate and optimization strategy, the architectural differences (depth and neuron count) heavily influence how these hyperparameters are tuned. The Neural ODE's deeper network requires more careful handling of learning rates and optimization techniques, with a need for more iterations of fine-tuning through BFGS. On the other hand, the shallow UDE architecture benefits from a simpler, faster optimization process, with fewer parameters and more stable training using RMSProp. 
\textbf{Overall, the UDE model shows better stability and generalization due to its less complex structure, and is more forgiving in terms of hyperparameter sensitivity}.

The following comparisons showcase the impact of hidden units, activation functions, and step size on the loss variation for both Neural ODE and UDE models.

\subsection{Forecasting with the Trained Models}

After training, the models are used to forecast the predator-prey populations beyond the training window, over the extended time span \( t \in [0, 20] \). The forecasting performance is evaluated by comparing the predictions with the true populations.

For the Neural ODE model, the neural network models the entire system dynamics, while for the UDE model, only the learned interaction terms are forecasted. Both models' forecasts are compared to assess whether retaining known physics (as in the UDE approach) provides an advantage over fully data-driven Neural ODEs.

\subsection{Model Validation and Performance}

To evaluate the model's forecasting accuracy, we calculate the root mean square error (RMSE) between the forecasted and actual prey and predator populations. Additionally, we introduce the concept of the \textbf{forecast breakdown point}, the point at which the model's predictions start to diverge significantly from the true values. This breakdown point is used to assess the robustness of UDEs versus Neural ODEs in long-term forecasting tasks.

Through this methodology, we demonstrate how UDEs can effectively learn unknown interaction terms while retaining known system dynamics, providing a flexible yet interpretable framework for ecological modeling.

\section{Results}

In this section, we analyze the performance of the Neural ODE and UDE models in terms of their ability to forecast the population dynamics of the Lotka-Volterra system as the amount of training data is progressively reduced. The goal is to identify the forecast breakdown point, where the model’s ability to predict future population dynamics deteriorates. We assess different cases of training data percentages for both models.

\subsection{Neural ODE Forecast Breakdown}

The Neural ODE model was trained on varying amounts of the dataset, and its forecasting performance was evaluated on the remaining unseen data. The following cases illustrate the breakdown points as the training data is reduced:

\subsubsection{Case 1: Training Neural ODE with 90\% data and forecasting on the remaining 10\%}  
In this case, the Neural ODE model was able to forecast with high accuracy, closely matching the ground truth for both prey and predator populations. (See Fig.~\ref{fig:ode_90_10})

\subsubsection{Case 2: Training Neural ODE with 50\% data and forecasting on the remaining 50\%}  
The model was still able to capture the underlying dynamics with very good accuracy (See Fig.~\ref{fig:ode_50_50})

\subsubsection{Case 3: Training Neural ODE with 40\% data and forecasting on the remaining 60\%}  
The model was still able to capture the underlying dynamics although slight perturbations start to appear (See Fig.~\ref{fig:ode_40_60})

\subsubsection{Case 4: Training Neural ODE with 35\% data and forecasting on the remaining 65\%}  
\begin{figure}[h]
    \centering
    \begin{subfigure}{0.45\textwidth}
        \centering
        \includegraphics[width=\linewidth]{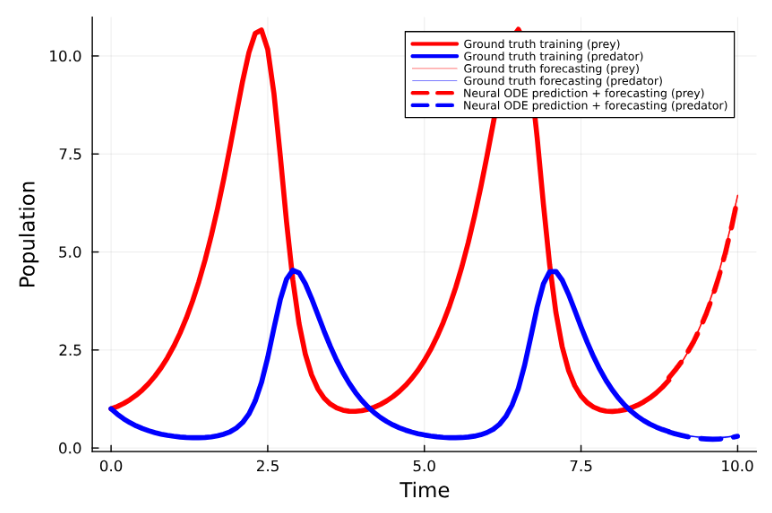}
        \caption{Neural ODE forecasting with 90\% training data and 10\% forecasting.}
        \label{fig:ode_90_10}
    \end{subfigure}
    \hfill
    \begin{subfigure}{0.45\textwidth}
        \centering
        \includegraphics[width=\linewidth]{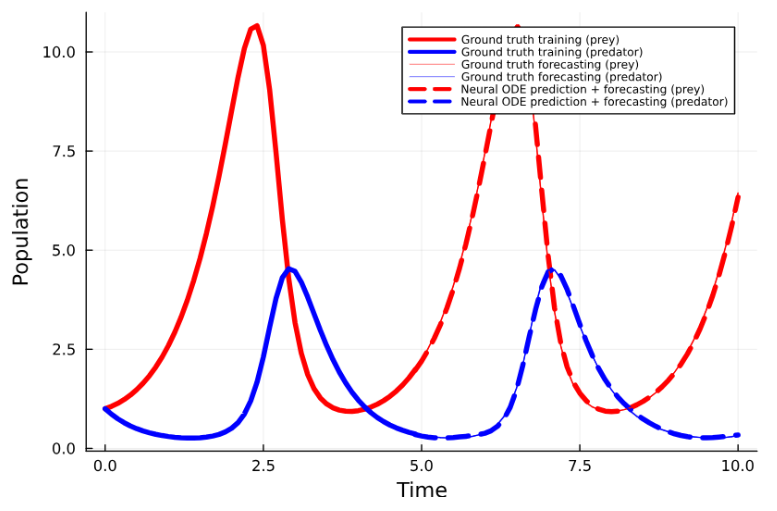}
        \caption{Neural ODE forecasting with 50\% training data and 50\% forecasting.}
        \label{fig:ode_50_50}
    \end{subfigure}
    
    \vspace{0.5cm}
    
    \begin{subfigure}{0.45\textwidth}
        \centering
        \includegraphics[width=\linewidth]{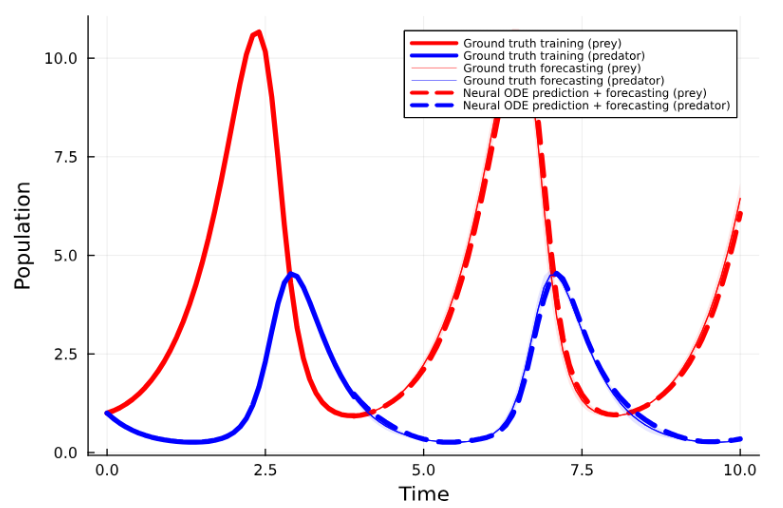}
        \caption{Neural ODE forecasting with 40\% training data and 60\% forecasting.}
        \label{fig:ode_40_60}
    \end{subfigure}
    \hfill
    \begin{subfigure}{0.45\textwidth}
        \centering
        \includegraphics[width=\linewidth]{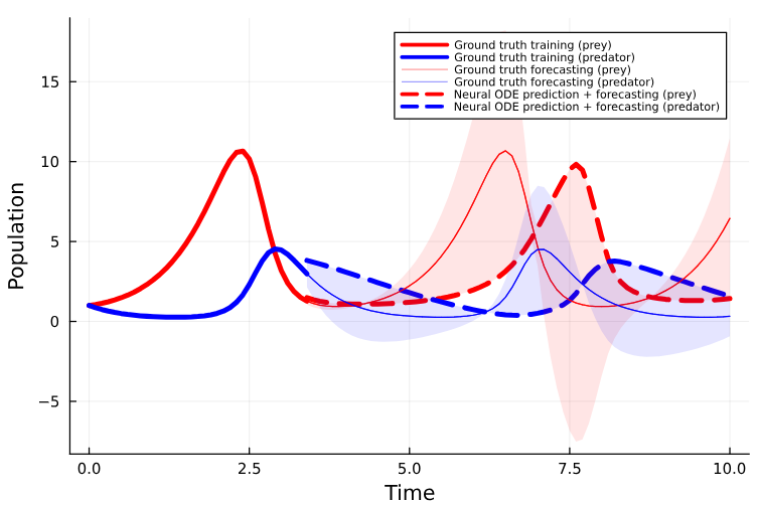}
        \caption{Neural ODE breakdown with 35\% training data and 65\% forecasting.}
        \label{fig:ode_35_65}
    \end{subfigure}
    
    \caption{Neural ODE Forecast Breakdown: Forecasts based on varying training data percentages.}
    \label{fig:neural_ode_forecast_breakdown}
\end{figure}

When trained on 35\% of the data, the Neural ODE model completely broke down. The forecast deviated significantly from the true population dynamics, indicating that the model could no longer capture the underlying system behavior accurately. (See Fig.~\ref{fig:ode_35_65})

\subsection{UDE Forecast Breakdown}

The UDE model was also trained on varying amounts of data to evaluate its performance under different training regimes.\\ We started from 35\% training data where Neural ODE model started to fail miserably. Notably, the UDE model demonstrated much better performance when trained on small portions of the dataset. Even when the training data was reduced to 35\%, the UDE model continued to forecast the population dynamics accurately, unlike the Neural ODE.

\subsubsection{Case 1: Training UDE with 35\% data and forecasting on the remaining 65\%}  
Unlike the Neural ODE, the UDE model was able to capture the underlying dynamics with much greater accuracy, even when trained on only 35\% of the data. This highlights the UDE model's robustness and ability to generalize with limited data. (See Fig.~\ref{fig:ude_forecast_35_65})

\subsubsection{Case 2: Partial Breakdown at 31\% Training Data}

The UDE model began to show signs of breakdown at 31\% training data, though it still captured the underlying dynamics reasonably well. This partial breakdown highlights the limitations of the UDE model when trained on extremely small data samples. (See Fig.~\ref{fig:ude_forecast_31_69})

\subsubsection{Case 3: Complete Breakdown at 30\% Training Data}

\begin{figure}[h]
    \centering
    \begin{subfigure}{0.45\textwidth}
        \centering
        \includegraphics[width=\linewidth]{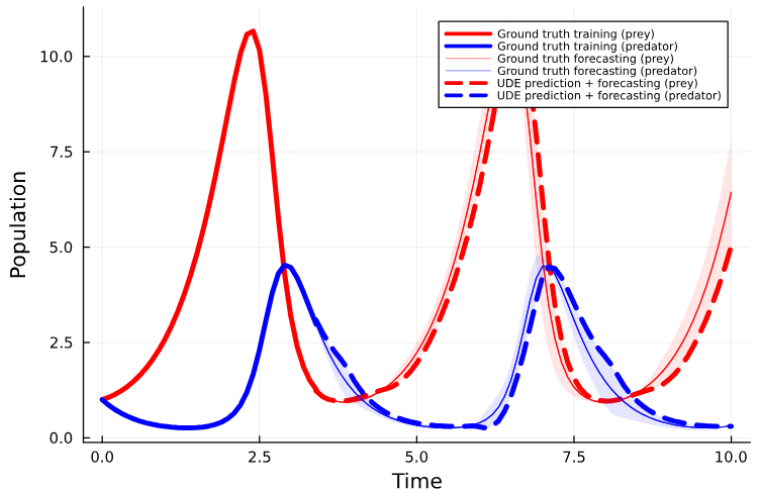}
        \caption{UDE forecasting with 35\% training data and 65\% forecasting.}
        \label{fig:ude_forecast_35_65}
    \end{subfigure}
    \hfill
    \begin{subfigure}{0.45\textwidth}
        \centering
        \includegraphics[width=\linewidth]{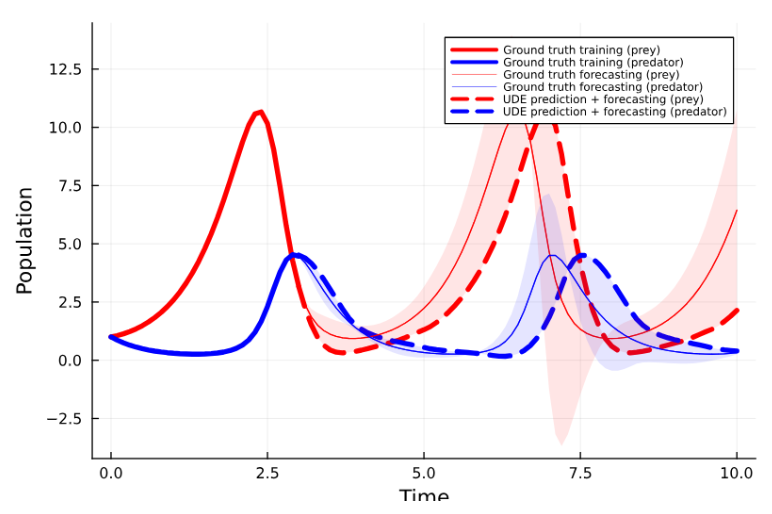}
        \caption{Partial breakdown in UDE with 31\% training data and 69\% forecasting.}
        \label{fig:ude_forecast_31_69}
    \end{subfigure}
    
    \vspace{0.5cm}
    
    \begin{subfigure}{0.45\textwidth}
        \centering
        \includegraphics[width=\linewidth]{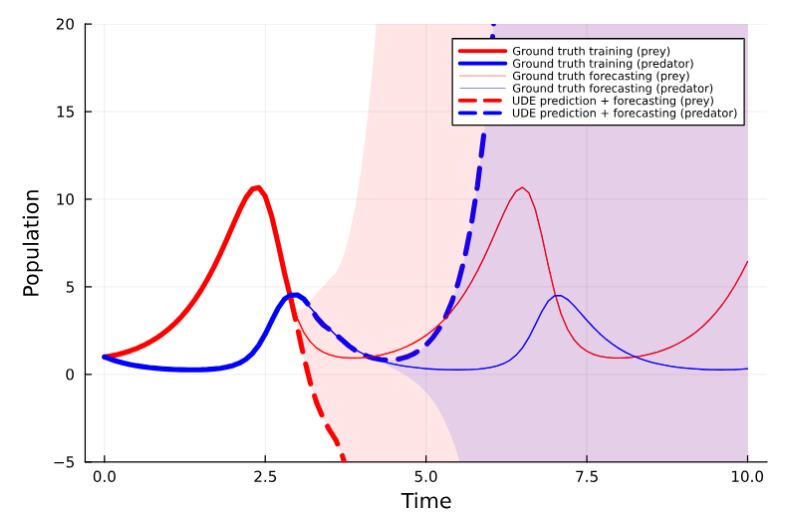}
        \caption{Complete breakdown in UDE with 30\% training data and 70\% forecasting.}
        \label{fig:ude_forecast_30_70}
    \end{subfigure}
    
    \caption{UDE Forecast Breakdown: (a) 35\% training data with 65\% forecasting, (b) Partial breakdown at 31\% training data, (c) Complete breakdown at 30\% training data.}
    \label{fig:ude_forecast_breakdown}
\end{figure}

Finally, we identified an edge case where the UDE model began to completely breakdown at 30\% training data. (See Fig.~\ref{fig:ude_forecast_31_69})

\subsection{Effect of Gaussian Noise on Performance}

To further test the robustness of Neural ODE and UDE, Gaussian noise with varying standard deviations was added to the training data. Specifically, noise was added with standard deviations of 0.05 and 0.3, representing low and high levels of noise, respectively.

\subsubsection{Performance with Mild Gaussian Noise (Standard Deviation = 0.05)}

When introducing Gaussian noise with a standard deviation of 0.05, both Neural ODE and UDE models showed resilience. The noise introduced only minor fluctuations in the data, and both models were able to learn the underlying dynamics without any significant degradation in performance.

\begin{itemize}
    \item \textbf{Neural ODE:} The model showed resilience and captured the underlying trend. It was able to capture both predator and prey data well. 
    
    \item \textbf{UDE:} UDE also demonstrated similar robustness under the same conditions. The model remained highly stable, maintaining accurate forecasts with little fluctuation in its predictions. 
\end{itemize}

    \begin{figure}[h]
        \centering
        \begin{subfigure}[b]{0.45\textwidth}
            \centering
            \includegraphics[width=\textwidth]{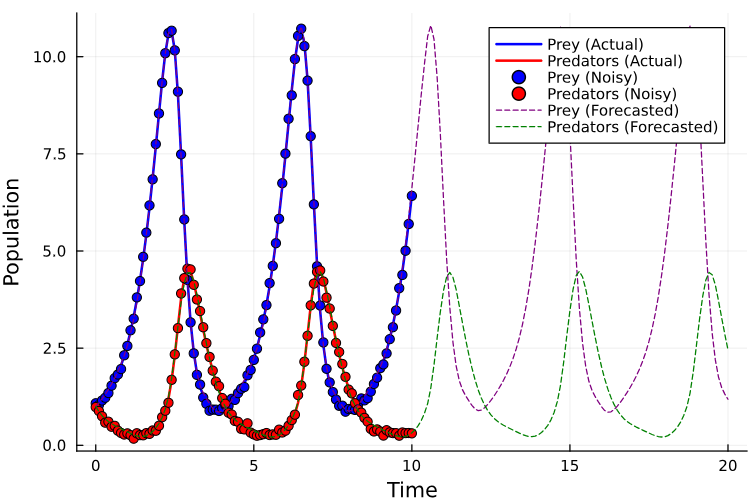}
            \caption{Neural ODE with 0.05 Gaussian noise}
            \label{fig:neural_ode_noise}
        \end{subfigure}
        \hfill
        \begin{subfigure}[b]{0.45\textwidth}
            \centering
            \includegraphics[width=\textwidth]{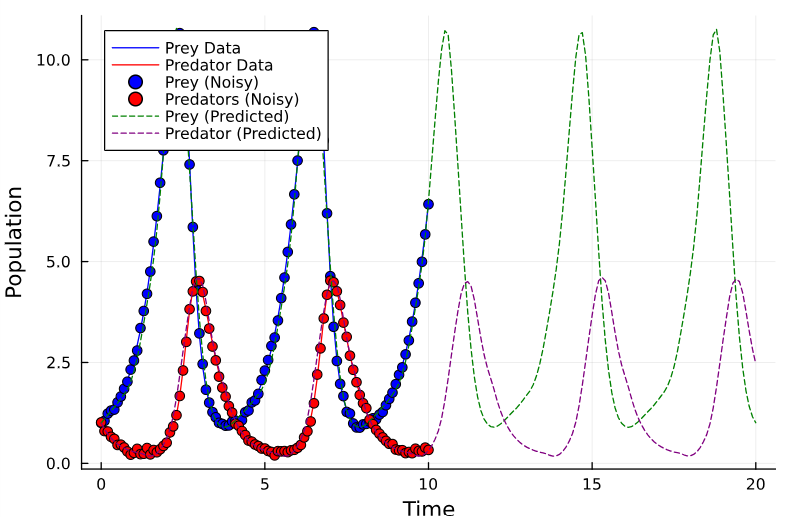}
            \caption{UDE with 0.05 Gaussian noise}
            \label{fig:ude_noise}
        \end{subfigure}
        \caption{Forecast capability comparison under 0.05 standard deviation Gaussian noise for Neural ODE (left) and UDE (right).}
        \label{fig:forecast_comparison_noise_0.05}
    \end{figure}

\subsubsection{Performance with High Gaussian Noise (Standard Deviation = 0.3)}

Introducing a higher level of Gaussian noise, with a standard deviation of 0.3, revealed a clear performance difference between Neural ODE and UDE. The added noise significantly distorted the data, making it much harder for the models to learn the dynamics accurately. However, UDE  outperformed Neural ODE:

\begin{itemize}
    \item \textbf{Neural ODE:} The performance of the Neural ODE model degraded notably. The loss function increased by over 25\%, and the model started to struggle with convergence during training. Forecast accuracy declined sharply, especially for predator data and for extended time horizons, where predictions diverged significantly from the true dynamics. This suggests that Neural ODE is highly sensitive to data corruption from noise and struggles with generalization under such conditions.
    
    \item \textbf{UDE:} On the other hand, UDE continued to perform relatively robustly despite the high noise levels. While the loss function increased by about 10\%, the model remained stable and produced reasonably accurate forecasts. UDE's ability to handle higher noise levels highlights its robustness, especially in real-world applications where data may often be noisy or imperfect.
\end{itemize}

        \begin{figure}[h]
        \centering
        \begin{subfigure}[b]{0.45\textwidth}
            \centering
            \includegraphics[width=\textwidth]{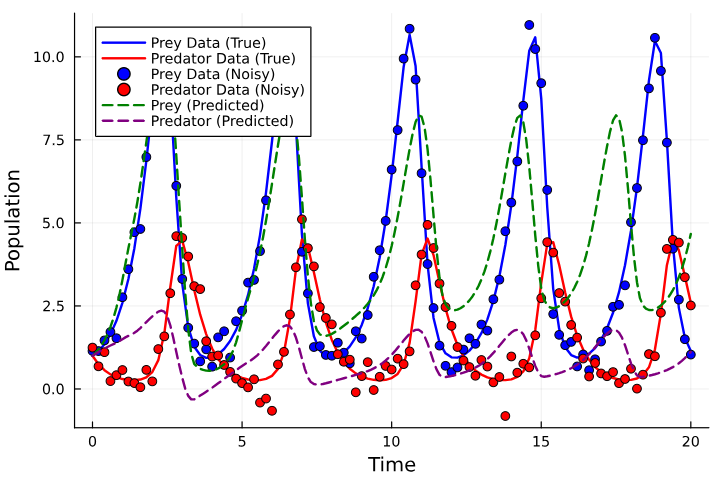}
            \caption{Neural ODE with 0.3 Gaussian noise}
            \label{fig:neural_ode_noise}
        \end{subfigure}
        \hfill
        \begin{subfigure}[b]{0.45\textwidth}
            \centering
            \includegraphics[width=\textwidth]{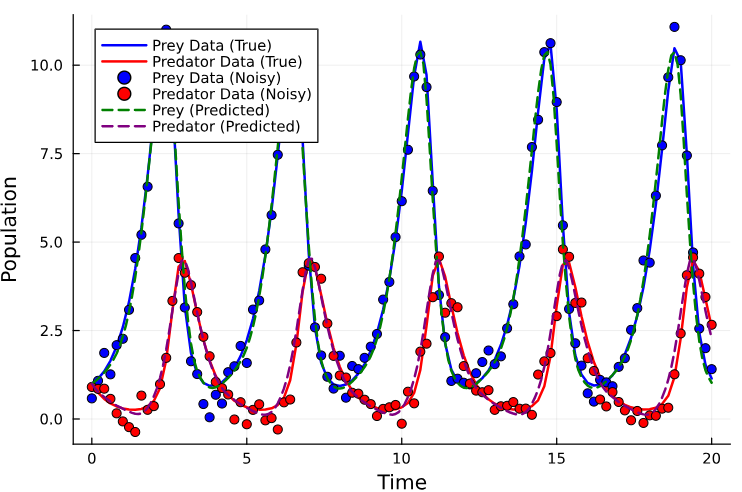}
            \caption{UDE with 0.3 Gaussian noise}
            \label{fig:ude_noise}
        \end{subfigure}
        \caption{Forecast capability comparison under 0.3 standard deviation Gaussian noise for Neural ODE (left) and UDE (right).}
        \label{fig:forecast_comparison_noise_0.3}
    \end{figure}

\section{Discussion and Conclusion}

In this work, we evaluated the performance of Neural Ordinary Differential Equations (Neural ODEs) and Universal Differential Equations (UDEs) in predicting the population dynamics of the Lotka-Volterra system. The primary focus was to identify the forecasting breakdown points as the amount of training data was progressively reduced, and to assess the effectiveness of each approach in terms of model architecture, data requirements, and generalization ability.

\subsection{Performance of Neural ODEs}

Neural ODEs offer a powerful framework for learning continuous dynamical systems directly from data. In our experiments, the Neural ODE model performed well when trained on a large percentage of the dataset (90\% or 50\%). The model successfully captured the underlying prey-predator interactions and provided accurate forecasts for future population dynamics.

However, as the amount of training data was reduced, the Neural ODE model began to show signs of deterioration. Specifically, the model started to break down when trained with less than 40\% of the data, with a complete failure observed at 35\% and below. This breakdown can be attributed to several factors:

\begin{itemize}
    \item \textbf{Model Complexity:} The Neural ODE architecture employed in this study was relatively deep, with three hidden layers of 100 neurons each. While this deep architecture provides expressive power, it also requires a large amount of data to generalize well. With less training data, the model struggled to learn the complex dynamics, leading to overfitting and poor forecasting performance.
    
    \item \textbf{Data Dependency:} Neural ODEs rely heavily on data for uncovering the complete system dynamics. As the amount of training data decreases, the model lacks the necessary information to learn the continuous trajectories of the system. This is particularly challenging for Neural ODEs, as they must approximate the full dynamics from scratch, requiring both a deep architecture and sufficient data.
\end{itemize}

\subsection{Performance of Universal Differential Equations (UDEs)}

In contrast to Neural ODEs, Universal Differential Equations (UDEs) performed significantly better in the same forecasting tasks, particularly when trained on smaller portions of the dataset. Even with as little as 35\% training data, the UDE model was able to capture the underlying dynamics of the Lotka-Volterra system with high accuracy. This robustness to data scarcity is one of the key advantages of UDEs.

Several factors contributed to the superior performance of UDEs:

\begin{itemize}
    \item \textbf{Partial Learning of Dynamics:} UDEs benefit from integrating known system dynamics (in this case, the Lotka-Volterra equations) with learned residual terms. Instead of learning the entire dynamics from scratch, the UDE model only learns the unknown or hard-to-model components of the system. This reduces the complexity of the learning task, allowing the model to generalize well even with less data. By leveraging prior knowledge, UDEs avoid overfitting and perform better under data constraints.
    
    \item \textbf{Shallow Neural Networks:} The UDE model used a much shallower architecture compared to the Neural ODE model—specifically, three hidden layers with only 10 neurons per layer. Despite this shallow network, the UDE model achieved better performance, highlighting the efficiency of the UDE framework. Shallow networks are not only computationally more efficient but also less prone to overfitting, particularly when trained on limited data. The ability to train shallow networks while maintaining high accuracy underscores the robustness of UDEs in capturing the system's behavior with fewer parameters.
    
    \item \textbf{Robustness to Data Scarcity:} One of the most significant advantages of UDEs is their robustness to data scarcity. While Neural ODEs showed a clear breakdown at 35\% training data, the UDE model continued to perform well. Even at 31\% training data, where some partial breakdown was observed, the model still captured the underlying dynamics reasonably well. This robustness is a result of combining known physics with machine learning, where the UDE framework augments the existing system dynamics with learned corrections. As a result, the model is less dependent on data and more capable of making accurate predictions with limited information.
    
    \item \textbf{Generalization Ability:} By leveraging the known Lotka-Volterra equations, UDEs focus on learning the specific corrections or interaction terms ($\beta$xy, $\gamma$xy) that may vary with data, rather than re-learning the entire dynamics. This allows the UDE model to generalize better across different time spans, even when training data is limited. The ability to extend the forecasting window while maintaining accuracy further demonstrates the advantage of UDEs over Neural ODEs in terms of generalization.

\end{itemize}

\subsection{Comparison of Computational Efficiency}

Another important factor to consider is computational efficiency. The Neural ODE model, with its deeper architecture, required more computational resources and time to converge, especially during the initial optimization phases. \\The UDE model, on the other hand, was not only more accurate with less data, but it was also faster to train due to its shallow network architecture. This efficiency is particularly valuable in scientific machine learning applications, where computational resources may be limited or expensive to scale.

\subsection{Conclusion}

In conclusion, while Neural ODEs are a powerful tool for modeling continuous dynamical systems, they are highly data-dependent and prone to breakdown when trained with insufficient data. Their deep architecture, while expressive, makes them less robust when data is limited. Universal Differential Equations (UDEs), on the other hand, provide a more robust, efficient, and accurate alternative. By leveraging known system dynamics and focusing on learning only the unknown components, UDEs perform better with less training data and shallower networks.

To summarize and conclude the key results, we present a comparison of Neural ODEs and UDEs in Table \ref{tab:summary_results}, which highlights the forecasting breakdown point, optimized hyperparameters, and the percentage decrease in loss during training.

\begin{table}[h]
    \centering
    \begin{tabular}{|l|c|c|}
        \hline
        \textbf{Criteria} & \textbf{Neural ODE} & \textbf{UDE} \\
        \hline
        \textbf{Forecasting Breakdown Point} & 35\% training data & 30\% training data \\
        \hline
        \textbf{Optimizers Used} & 
        \begin{tabular}[c]{@{}c@{}}
            Adam (LR = 0.001)\\
            BFGS for fine-tuning\\
            Adam for 500 iterations,\\
            BFGS for 100 iterations
        \end{tabular} & 
        \begin{tabular}[c]{@{}c@{}}
            Adam (LR = 0.001)\\
            RMSProp for fine-tuning\\
            Adam for 20,000 iterations,\\
            RMSProp for 5,000 iterations
        \end{tabular} \\
        \hline
        \textbf{Loss Decrease (\%)} & 
        \begin{tabular}[c]{@{}c@{}}
            85\% after Adam and BFGS\\
            Loss function: sum of squared differences
        \end{tabular} & 
        \begin{tabular}[c]{@{}c@{}}
            90\% after Adam and RMSProp\\
            Loss function: sum of squared differences
        \end{tabular} \\
        \hline
        \textbf{Neural Network Architecture} & 
        \begin{tabular}[c]{@{}c@{}}
            3 hidden layers with 100 neurons each\\
            Activation function: RBF\\
            Dense Network
        \end{tabular} & 
        \begin{tabular}[c]{@{}c@{}}
            3 hidden layers with 10 neurons each\\
            Activation function: ReLU\\
            Shallow network
        \end{tabular} \\
        \hline
        \textbf{Training Time (approx)} & 50 minutes & 20 minutes \\
        \hline
        \textbf{Parameter Count} & 
        \begin{tabular}[c]{@{}c@{}}
            $\sim$35,000 parameters\\
            (due to 100 neurons per layer)
        \end{tabular} & 
        \begin{tabular}[c]{@{}c@{}}
            $\sim$3,500 parameters\\
            (due to 10 neurons per layer)
        \end{tabular} \\
        \hline
        \textbf{Sensitivity to Data Scarcity} & High & Low \\
        \hline
        \textbf{Model Robustness on Small Data} & Poor & Strong \\
        \hline
        \textbf{Forecast Accuracy} & 
        \begin{tabular}[c]{@{}c@{}}
            Breaks down beyond\\
            35\% training data
        \end{tabular} & 
        \begin{tabular}[c]{@{}c@{}}
            Accurate forecasts even with\\
            30\% training data
        \end{tabular} \\
        \hline
        \textbf{Sensitivity to Noise } & 
        \begin{tabular}[c]{@{}c@{}}
            Unable to recover underlying trend\\
            with noise (Gaussian, $\sigma=0.3$)
        \end{tabular} & 
        \begin{tabular}[c]{@{}c@{}}
            Able to capture underlying trend\\
            despite noise (Gaussian, $\sigma=0.3$)
        \end{tabular} \\
        \hline
    \end{tabular}
    
    \caption{Summary of results comparing Neural ODEs and UDEs.}
    \label{tab:summary_results}
\end{table}

The table highlights several key differences between the two approaches:

\begin{itemize}
    \item \textbf{Forecasting Breakdown Point:} UDEs were able to maintain accuracy even with 30\% training data, while Neural ODEs completely broke down when trained with less than 35\%.
    \item \textbf{Optimized Hyperparameters:} Both models used the Adam optimizer with a learning rate of 0.001 initially, but UDEs switched to RMSProp for further fine-tuning, while Neural ODEs used BFGS for final optimization.
    \item \textbf{Loss Decrease:} UDEs achieved a higher percentage decrease in loss during training, indicating more effective optimization with fewer iterations compared to Neural ODEs.
    \item \textbf{Neural Network Architecture:} UDEs achieved superior performance using a much shallower network (3 layers with 10 neurons per layer) compared to Neural ODEs (3 layers with 100 neurons per layer). The UDE model used a ReLU activation function, which is computationally cheaper and easier to optimize, while the Neural ODE used RBF, which is better for continuous modeling but adds computational complexity.
    \item \textbf{Training Time:} UDEs were computationally more efficient, requiring only 20 minutes to train, while Neural ODEs took approximately 50 minutes due to their deeper architecture and more complex optimization process.
    \item \textbf{Parameter Count:} The Neural ODE model has an order of magnitude more parameters than the UDE model, leading to higher computational requirements and a higher risk of overfitting when data is limited.
    \item \textbf{Model Robustness:} UDEs demonstrated stronger robustness to data scarcity, showing better generalization when trained on as little as 30\% of the dataset.
    \item \textbf{Sensitivity to Noise:} UDEs were able to capture the underlying trend even when Gaussian noise with a standard deviation of 0.3 was introduced into the data. In contrast, Neural ODEs struggled to recover the underlying trend in the presence of this noise, highlighting UDEs' superior robustness to noisy, real-world data.
\end{itemize}

The key advantages of UDEs include their ability to generalize well, their robustness to data scarcity, noise and their computational efficiency. This makes UDEs particularly well-suited for scientific machine learning tasks where prior knowledge of the system is available, and training data may be scarce. Overall, UDEs offer a compelling framework for combining machine learning with differential equation modeling, outperforming Neural ODEs in scenarios where data is limited or noisy.

\section*{Limitations and Future Work}
While our approach using UDEs offers several advantages, such as robustness to noise, data scarcity, and computational efficiency, there are certain limitations that must be addressed for broader applicability. One significant challenge is the training of SciML frameworks on systems with a large volume of data points. As the dataset size increases, the computational cost and memory requirements also grow substantially, making it difficult to train models using higher-order differential equations or models with complex interactions, where each iteration requires significant computational power and memory for solving the differential equations accurately.

Furthermore, while UDEs can incorporate prior knowledge about the system, the process of defining the right combination of neural network architectures and differential equations remains non-trivial. It requires a deep understanding of both the physical system and the appropriate modeling techniques, making the framework less accessible for practitioners who lack domain expertise. 

For future work, we see opportunities to enhance the scalability of UDEs through the development of more efficient solvers and optimization algorithms, potentially leveraging parallel computing and GPU acceleration. 

We also plan to explore the application of UDEs to a wider range of real-world biological and ecological systems, such as multi-species interactions, time-series \cite{kidger2020neural}, graphs \cite{xia2021graph}, sequential data \cite{lechner2019designing}, ecosystem dynamics, and epidemiological models. These systems often suffer from a scarcity of high-quality training data and exhibit complex, non-linear dynamics, making them ideal candidates for the advantages of UDEs. In particular, our future studies will aim to quantify the robustness of UDEs across different types of noise and data irregularities, providing a more comprehensive understanding of their capabilities in real-world applications.

Moreover, integrating UDEs with other advanced modeling techniques, such as graph neural networks and stochastic differential equations \cite{yildiz2019ode2vae}, could further extend their applicability to more intricate systems with spatial and temporal variability. This approach would allow us to capture not only the temporal evolution of systems but also the spatial interactions between different components, offering a holistic perspective on the dynamics of complex natural phenomena.

In summary, while the current study demonstrates the potential of UDEs in modeling and forecasting ecological systems like the Lotka-Volterra predator-prey model, addressing these challenges will be critical to fully realizing the potential of UDEs in scientific machine learning. By focusing on scalability, ease of use, and application to diverse real-world problems, we aim to make UDEs a versatile tool for scientists and researchers in understanding and predicting complex systems.

\section*{Acknowledgements}

I would like to express my deepest gratitude to Dr Raj Abhijit Dandekar, Dr Rajat Dandekar, and Dr Sreedath Panat for their invaluable guidance, insightful feedback and support during the course of this work. Special thanks to Dr Raj Abhijit Dandekar for his keen observations and for providing crucial guidance that greatly enhanced the direction and quality of the research.

\bibliographystyle{unsrt}  
\bibliography{references}  

\begin{thebibliography}{10}

\bibitem{ref1}
Ricky T.~Q. Chen, Yulia Rubanova, Jesse Bettencourt, and David Duvenaud.
\newblock Neural ordinary differential equations.
\newblock In {\em Advances in Neural Information Processing Systems}, 2018.

\bibitem{PINNeqn}
M.~Raissi, P.~Perdikaris, and G.~E. Karniadakis.
\newblock Physics-informed neural networks: A deep learning framework for solving forward and inverse problems involving nonlinear partial differential equations.
\newblock {\em Journal of Computational Physics}, 378, 2019.

\bibitem{rackauckas2020universal}
Chris Rackauckas, Yingbo Ma, Julius Martensen, Collin Warner, Kirill Zubov, Rohit Supekar, Dominic Skinner, and Alan Edelman.
\newblock Universal differential equations for scientific machine learning.
\newblock {\em arXiv preprint arXiv:2001.04385}, 2020.

\bibitem{rackauckas2017differentialequations}
Chris Rackauckas and Qing Nie.
\newblock Differentialequations.jl – a performant and feature-rich ecosystem for solving differential equations in julia.
\newblock {\em Journal of Open Research Software}, 5(1), 2017.

\bibitem{ma2021comparison}
Yingbo Ma, Vaibhav Dixit, and SciML et~al.
\newblock A comparison of automatic differentiation and continuous sensitivity analysis for derivatives of differential equation solutions.
\newblock In {\em Advances in Neural Information Processing Systems}, volume~34, 2021.

\bibitem{dandekar2020quantifying}
Raj Dandekar and George Barbastathis.
\newblock Quantifying the effect of quarantine control in covid-19 infectious spread using machine learning.
\newblock {\em Chaos, Solitons \& Fractals}, 2020.

\bibitem{toth2020hamiltonian}
Peter Toth, Bianca Dumitrescu, and David Duvenaud.
\newblock Hamiltonian neural networks for solving differential equations.
\newblock In {\em Advances in Neural Information Processing Systems}, volume~33, 2020.

\bibitem{rackauckas2019scalable}
Chris Rackauckas, Yingbo Ma, and Alan Edelman.
\newblock A scalable solution to parallelizing scientific machine learning on supercomputers.
\newblock In {\em IEEE/ACM Supercomputing Conference}, 2019.

\bibitem{kidger2020neural}
Patrick Kidger, James Foster, Ricky~T.Q. Chen, and Terry Lyons.
\newblock Neural controlled differential equations for irregular time series.
\newblock In {\em Advances in Neural Information Processing Systems}, 2020.

\bibitem{hinton2012neural}
Geoffrey Hinton, Nitish Srivastava, and Kevin Swersky.
\newblock Neural networks for machine learning, 2012.
\newblock Coursera Lecture Notes.

\bibitem{goodfellow2016deep}
Ian Goodfellow, Yoshua Bengio, and Aaron Courville.
\newblock {\em Deep learning}.
\newblock MIT Press, 2016.

\bibitem{nair2010rectified}
Vinod Nair and Geoffrey~E. Hinton.
\newblock Rectified linear units improve restricted boltzmann machines.
\newblock In {\em Proceedings of the 27th International Conference on Machine Learning (ICML-10)}, 2010.

\bibitem{glorot2011deep}
Xavier Glorot, Antoine Bordes, and Yoshua Bengio.
\newblock Deep sparse rectifier neural networks.
\newblock In {\em Proceedings of the Fourteenth International Conference on Artificial Intelligence and Statistics}, 2011.

\bibitem{maas2013rectifier}
Andrew Maas, Awni Hannun, and Andrew Ng.
\newblock Rectifier nonlinearities improve neural network acoustic models.
\newblock In {\em Proceedings of the 30th International Conference on Machine Learning}, 2013.

\bibitem{kingma2014adam}
Diederik~P. Kingma and Jimmy Ba.
\newblock Adam: A method for stochastic optimization.
\newblock {\em arXiv preprint arXiv:1412.6980}, 2014.

\bibitem{wilson2017marginal}
Ashia~C. Wilson, Rebecca Roelofs, Mitchell Stern, Nati Srebro, and Benjamin Recht.
\newblock The marginal value of adaptive gradient methods in machine learning.
\newblock In {\em Advances in Neural Information Processing Systems}, 2017.

\bibitem{choi2019empirical}
Dongchan Choi, Roger Grosse, and James Martens.
\newblock On empirical comparisons of optimizers for deep learning.
\newblock {\em arXiv preprint arXiv:1910.05446}, 2019.

\bibitem{loshchilov2017fixing}
Ilya Loshchilov and Frank Hutter.
\newblock Fixing weight decay regularization in adam.
\newblock {\em arXiv preprint arXiv:1711.05101}, 2017.

\bibitem{powell2009bfgs}
Michael~J.D. Powell.
\newblock The broyden–fletcher–goldfarb–shanno (bfgs) optimization algorithm.
\newblock In {\em Encyclopedia of Optimization}. Springer, 2009.

\bibitem{liu1989limited}
Dong~C. Liu and Jorge Nocedal.
\newblock On the limited memory bfgs method for large scale optimization.
\newblock {\em Mathematical Programming}, 45(1-3), 1989.

\bibitem{shanno1970conditioning}
David~F. Shanno.
\newblock Conditioning of quasi-newton methods for function minimization.
\newblock {\em Mathematics of Computation}, 24(111), 1970.

\bibitem{tieleman2012rmsprop}
Tijmen Tieleman and Geoffrey Hinton.
\newblock Rmsprop: Divide the gradient by a running average of its recent magnitude, 2012.
\newblock Coursera Lecture Notes.

\bibitem{xia2021graph}
Yitian Xia, Tianfu He, and Rajat Talak.
\newblock Graph neural ordinary differential equations.
\newblock {\em IEEE Transactions on Neural Networks and Learning Systems}, 2021.

\bibitem{lechner2019designing}
Mathias Lechner and Ramin Hasani.
\newblock Designing neural ordinary differential equations for continuous learning.
\newblock In {\em Advances in Neural Information Processing Systems (NeurIPS)}, 2019.

\bibitem{yildiz2019ode2vae}
Cagatay Yildiz, Maximilian Golub, and Patrick van~der Smagt.
\newblock Ode2vae: Deep generative second-order odes with bayesian neural networks.
\newblock In {\em Advances in Neural Information Processing Systems}, 2019.

\end{thebibliography}

\end{document}